\documentclass[11pt]{article}
\usepackage[preprint]{acl} 

\usepackage{times}
\usepackage{latexsym}
\usepackage{inconsolata}

\usepackage{amsmath,amssymb}
\usepackage{booktabs}
\usepackage{siunitx}
\usepackage{graphicx}
\usepackage{algpseudocode}
\usepackage{microtype}
\usepackage{url}
\usepackage{hyperref}
\usepackage{algorithm}
\usepackage{xspace}
\usepackage{float}
\usepackage{placeins}

\usepackage{xcolor}
\definecolor{darkblue}{HTML}{000099}
\hypersetup{
    colorlinks=true,
    linkcolor=darkblue,
    filecolor=darkblue,      
    urlcolor=darkblue,
    citecolor=darkblue,
}

\usepackage{enumitem}

\usepackage{subcaption}

\title{Distributed Interpretability and Control for Large Language Models}
\author{Dev Arpan Desai, Shaoyi Huang, Zining Zhu\\
  \texttt{\{ddesai4,shuang59,zzhu41\}@stevens.edu}}

\begin{document}
\maketitle

\begin{abstract}
Large language models that require multiple GPU cards to host are usually the most capable models. It is necessary to understand and steer these models, but the current technologies do not support the interpretability and steering of these models in the multi-GPU setting as well as the single-GPU setting. We present a practical implementation of activation-level interpretability (logit lens) and steering (steering vector) that scales up to multi-GPU language models. Our system implements design choices that reduce the activation memory by up to 7x and increase the throughput by up to 41x compared to a baseline on identical hardware. We demonstrate the method across LLaMA-3.1 (8B, 70B) and Qwen-3 (4B, 14B, 32B), sustaining 20–100 tokens/s while collecting full layer-wise activation trajectories for sequences of 1,500 tokens. Using label-position steering vectors injected post-LayerNorm, we show controllable, monotonic shifts in model outputs with a mean steerability slope of 0.702 across evaluated datasets, without fine-tuning or additional forward passes. We release detailed benchmarks, ablations, and a reproducible instrumentation recipe to enable practical interpretability and real-time behavioral control for frontier LLMs at \url{https://github.com/Devdesai1901/LogitLense}.

\end{abstract}

\section{Introduction}

Large language models (LLMs) now drive applications in scientific analysis, healthcare, law, and interactive decision-making, increasing the need to understand \emph{why} a model produces a specific output and how we can reliably influence its behavior. As models scale into the tens of billions of parameters, we increasingly require interpretability tools that expose intermediate computations and enable targeted interventions during inference. Yet achieving visibility at this scale remains difficult. Interpretability methods must surface internal structure with fine granularity while preserving the throughput and memory constraints required for real-world deployment. Today, the models for which we most need interpretability are unfortunately the models for which interpretability remains least feasible.

Existing activation-level interpretability techniques including the logit lens and its variants and activation steering approaches provide conceptual insight but fail to scale to contemporary 10B+ parameter models. Prior implementations often (1) require multiple forward passes, (2) no KV caching, (3) project every intermediate activation into the full vocabulary, or (4) assume single-GPU execution. These design choices inflate memory usage, increase latency, and frequently trigger out-of-memory (OOM) failures, even on mid-sized models. As a result, activation-level interpretability and steering for large language models is effectively not possible.

In this work, we introduce a \emph{distributed, single-pass framework} for activation-level interpretability and behavioral steering that runs at native inference speed on models up to 70B parameters. We integrate instrumentation \emph{inside} the tensor-parallel inference path so we can capture activations during standard autoregressive decoding without extra forward passes, without interfering with KV caching, and without materializing full vocabulary logits. We instead record compact hidden-state slices for each token and perform a single batched LM-head projection only after decoding finishes. This design keeps memory proportional to the hidden dimension rather than vocabulary size, enabling full-layer, long-sequence analyses under fixed per-GPU budgets. Empirically, we reduce activation memory by up to 
7x and increase throughput by up to  41x compared to a reimplemented LogitLens4LLMs ~\cite{wang2025logitlens4llms} baseline on identical hardware.

We also show that this same architecture supports real-time \emph{behavioral steering}. We compute steering vectors from base and target forward activations and inject them post-LayerNorm inside wrapped transformer layers. Because we inject inside the existing tensor-parallel forward path, steering preserves single-pass execution and introduces no extra compute. Across LLaMA-3.1 (8B, 70B) and Qwen3 (4B, 14B, 32B), we observe monotonic, controllable output shifts with a mean steerability slope of  0.702, demonstrating that activation-level interventions remain reliable at frontier scale without fine-tuning or throughput degradation.

Prior interpretability methods such as logit lens and activation steering are developed under single-device or offline assumptions and are fundamentally incompatible with tensor-parallel, KV-cached inference used by modern large language models. In such settings, hidden states are sharded across devices and discarded incrementally during generation, making layer-wise decoding or intervention infeasible without additional forward passes or parameter modification. We address this gap by introducing a unified execution model that integrates interpretability and steering directly into tensor-parallel inference via single-pass activation capture and deferred projection.

\paragraph{Contributions}
We summarize our contributions below:

\begin{itemize}[nosep,leftmargin=*]
\item \textbf{Tensor-parallel interpretability.}
We introduce a tensor-parallel inference architecture that enables large language models (LLaMA, Qwen) to be efficiently analyzed and controlled across multiple GPUs. By integrating logit-lens analysis and activation steering directly into tensor-parallel execution, our approach supports full-layer interpretability during standard inference, enabling scalable analysis up to 70B  models at native resolution without centralizing model weights or activations.

\item \textbf{Memory-efficient, single-pass instrumentation.}  
We combine (i) single-pass activation capture, (ii) deferred vocabulary projection, and (iii) non-redundant hidden-state logging. This design reduces activation footprint by up to \(7\times\) and sustains 20–100 tokens/s while tracing 1,500-token sequences across all layers of LLaMA and Qwen models. 

\item \textbf{Scalable steering vectors with real-time control.}  
We implement post-LayerNorm steering inside the distributed forward path, achieving stable, monotonic behavioral modifications with no additional passes and no loss in throughput. Steering remains effective across all tested model scales. 

\end{itemize}

Together, these contributions close a practical gap: we make full-layer, token-level interpretability and activation-level control feasible upto 70B parameter models using commodity multi-GPU hardware, without slowing generation or exceeding memory budgets. Our results point toward unified interpretability-and-control systems suitable for both research and production.

\paragraph{Roadmap}
Section~\ref{sec:related_work} reviews interpretability and steering methods.
Section~\ref{sec:method} presents our distributed, single-pass architecture.
Section~\ref{sec:experiments} reports scalability and steering results across five models.
Section~\ref{sec:conclusion} discusses limitations and future directions.

\newcommand{\baseline}{LL4L--TP}
\newcommand{\ours}{SP--TP} 

\section{Related Work}
\label{sec:related_work}
\subsection{Logit Lens and Tuned Lens Methods}

The \emph{logit lens}, introduced by Nostalgebraist~\cite{nostalgebraist2020logitlens}, projects intermediate residual-stream activations through the output embedding to expose how token predictions evolve across layers. While it reveals a progression from diffuse to sharpened distributions, it suffers from \emph{basis drift}, leading to misaligned intermediate predictions. Belrose et al.~\cite{belrose2025tunedlens} addressed this limitation with the \emph{tuned lens}, learning a lightweight per-layer linear correction that substantially improves alignment between intermediate logits and final model outputs.

Subsequent tooling such as \textsc{LogitLens4LLMs}~\cite{wang2025logitlens4llms} extended these ideas to modern architectures (e.g., LLaMA and Falcon), adding improved visualization pipelines and projection-fidelity metrics. However, existing logit-lens implementations typically rely on multi-pass decoding or single-GPU execution, restricting practical applicability to models below roughly 10B parameters.

\begin{figure*}[ht]
\centering
\includegraphics[
  width=\linewidth,
  trim={1.0cm 2cm 1.0cm 3cm},
  clip
]{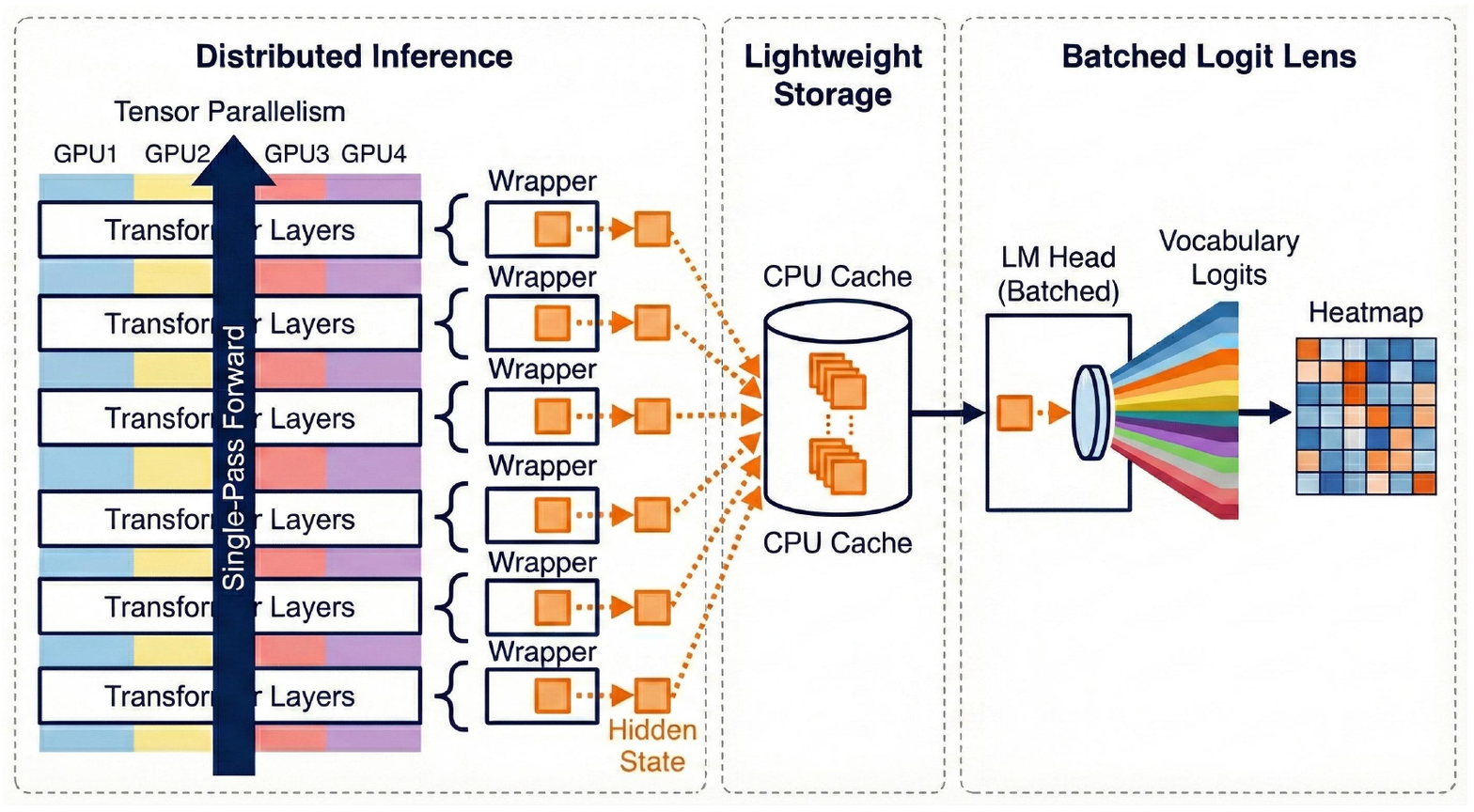}
\caption{
\textbf{Distributed single-pass logit-lens architecture.}
Selected transformer layers are instrumented with lightweight wrappers
that expose post-attention activations under tensor-parallel execution.
Hidden states for the most recent token are recorded during a single KV-cached forward pass.
After generation, all captured activations are decoded in batch via the final normalization layer
and shared LM head, enabling scalable per-token, per-layer analysis without additional forward passes.
}
\label{fig:arch}
\end{figure*}

\subsection{Steering Vectors and Activation Interventions}

Steering vectors provide a complementary mechanism for controlling model behavior by directly modifying hidden activations rather than reading them. \citet{turner2023activationengineering} showed that a simple direction computed from positive versus negative examples can reliably steer attributes such as sentiment or topic without retraining. Compared to gradient-based approaches such as PPLM, steering vectors are lightweight, reusable, and state-based.

Recent work has focused on robustness and scalability. \cite{tan2024analyzinggeneralizationreliabilitysteering} systematically evaluated steering vectors across models and layers, showing that effectiveness varies substantially with architecture and injection depth. Empirical demonstrations on deployed models include \cite{gunton2024vectorsteering}, who showed that mean-direction steering can effectively guide generative models toward improved reasoning behavior. Other work explores reducing interference: \cite{konen2024stylevectors} introduced style vectors for controlled stylistic shifts, while \cite{stoehr2024activationscaling} proposed sparse, layer-selective activation scaling to minimize side effects.

Open-source libraries such as \texttt{llm-steer} \citep{mihaiii2025llmsteer} and \texttt{steering-vectors} \citep{steeringvectors_lib} provide practical interfaces for constructing and applying steering vectors, though they do not yet address large-scale, tensor-parallel deployment.

\subsection{Engineering and Systems for Interpretability at Scale}

Several systems aim to support activation-level interpretability in large transformers, but most remain limited by model scale or execution constraints. Early pipelines for GPT-2  relied on per-layer projection or extensive activation logging, making them prohibitively expensive for larger models. More recent inference engines and tracing frameworks enable activation logging but do not support full residual-stream decoding or multi-layer steering during generation.

Crucially, prior systems do not jointly support \emph{single-pass activation capture}, \emph{deferred batched projection}, and \emph{multi-layer activation intervention} while preserving tensor-parallel throughput and KV-cache semantics. As a result, interpretability and steering methods have remained largely incompatible with realistic multi-GPU inference for 10B+ parameter models.

\section{System Architecture and Method}
\label{sec:method}

\subsection{Architecture Overview}
\label{sec:arch_overview}

We operationalize the design principles introduced earlier single-pass activation capture,
tensor-parallel execution, and deferred vocabulary projection into a three stage architecture
illustrated in Fig.~\ref{fig:arch}:
(1) distributed model initialization with selective layer instrumentation,
(2) single-pass autoregressive generation with activation capture,
and (3) batched post-hoc decoding and serialization.

\paragraph{Stage 1: Distributed initialization and layer instrumentation.}
We initialize the tokenizer and base language model under a tensor-parallel inference engine,
partitioning weights across all available GPUs.
A configurable subset of transformer layers is instrumented by replacing the original block
with a \emph{Block Output Wrapper}.
Each wrapper exposes post-attention activations from three locations:
(i) the attention mechanism output,
(ii) the MLP output,
and (iii) the block (residual-stream) output.
The model’s final normalization layer and shared LM head remain unmodified and are reused during post-hoc decoding.

\paragraph{Stage 2: Single-pass generation with activation capture.}
Generation proceeds via standard autoregressive decoding with key-value caching enabled.
Because instrumentation occurs directly within the forward path,
no additional forward passes are required.
At each decoding step and for each wrapped layer,
we record only the hidden-state slice corresponding to the most recently generated token.

Concretely, for a wrapped layer $\ell$ and activation type $c$,
the recorded tensor has shape $[1,1,d]$ (batch size $1$, single token, hidden dimension $d$).
These slices are appended during generation.
Across a generation of length $T$, each wrapped layer therefore accumulates
an activation trajectory of shape $[1,T,d]$ per activation type.
All capture occurs within the same tensor-parallel, KV-cached forward pass as standard inference.

\paragraph{Stage 3: Batched decoding and serialization.}
After generation completes, the per-token activation slices are concatenated
into full hidden-state trajectories.
For each wrapped layer and activation type, we apply the model’s final normalization
followed by a batched projection through the shared LM head,
yielding vocabulary logits for all tokens simultaneously.
Rather than retaining full $|V|$-dimensional distributions,
we extract only the top-$k$ candidates and their probabilities per token.
The resulting outputs are serialized in a JSON format,
indexed by layer and activation type, and consumed by downstream visualization
and analysis tools.
This design ensures that memory usage scales with sequence length and hidden dimension,
rather than vocabulary size.

\subsection{Single-pass Capture and Batched Decoding Algorithm}

\algnewcommand\algorithmicinput{\textbf{Input:}}
\algnewcommand\Input{\item[\algorithmicinput]}
\begin{algorithm}[h]
\caption{Single-pass activation capture with batched logit-lens decoding}
\begin{algorithmic}[1]
    \Require Model $\mathcal{M}$, Tokenizer $\tau$, Wrappers $\mathcal{L}$, Types $\mathcal{C}$
    \State Initialize DeepSpeed inference with tensor parallelism
    \State Instrument layers $\ell \in \mathcal{L}$ with Block Output Wrappers
    \State \textbf{Phase 1: Generation \& Capture}
    \State Run autoregressive generation ($t=1 \dots T$) with KV-cache:
    \State \quad Wrappers record slice $h \in \mathbb{R}^{1 \times 1 \times d}$ at step $t$
    \State \textbf{Phase 2: Post-hoc Decoding}
    \For{$\ell \in \mathcal{L}, c \in \mathcal{C}$}
        \State Concatenate slices to form $H^{(\ell,c)} \in \mathbb{R}^{1 \times T \times d}$
        \State $Z \gets \text{LMHead}(\text{LN}(H^{(\ell,c)}))$ \Comment{Batched Projection}
        \State Extract top-$k$ and $\text{Softmax}(\text{top-}k)$
    \EndFor
\end{algorithmic}
\end{algorithm}

\paragraph{Memory complexity.}
Storing captured activations scales as:
\[
\mathcal{O}\!\left(T \cdot d \cdot |\mathcal{L}| \cdot |\mathcal{C}|\right).
\]
For example, with $T=1500$ tokens, $d=8192$ vector embeddings size , $|\mathcal{L}|=80$ transformer layers and
C = number of activation types
this corresponds to approximately $9.8 \times 10^{8}$ elements.
Under bfloat16 precision (2 bytes per element), this is roughly $1.97$~GB per activation type
(or $\sim 5.9$~GB when storing attention, MLP, and block outputs),
which remains tractable on modern multi-GPU systems.

\subsection{Logit Lens Decoding}
\label{sec:logit_lens}

Our implementation of the logit lens operates as a post-hoc decoding step
over the hidden-state trajectories captured during generation.
By deferring projection into vocabulary space,
we avoid repeated $|V|$-dimensional computation during autoregressive decoding.

\paragraph{Batched projection.}
Given a hidden state $\mathbf{h}_{\ell,t} \in \mathbb{R}^{d}$ from layer $\ell$ at token position $t$, we compute vocabulary logits as
\begin{equation}
\mathbf{z}_{\ell,t}
=
W_{\text{out}} \, \mathrm{LN}\!\left(\mathbf{h}_{\ell,t}\right) + \mathbf{b},
\label{eq:logit-lens-equation}
\end{equation}
where $\mathrm{LN}$ denotes the model’s final normalization layer
and $W_{\text{out}}$ is the language model head of the model.

\paragraph{Top-$k$ extraction.}
For each token position, we extract the top-$k$ logits using an argsort operation.
The value of $k$ is specified by a user-controlled hyperparameter.
Probabilities are computed by applying a softmax over the top-$k$ logits only,
yielding a conditional distribution over the selected candidates rather than
a full-vocabulary normalization.
This choice substantially reduces memory and compute overhead
while preserving the relative ranking required for interpretability.

\subsection{Activation Steering}
\label{sec:activation_steering}

Our instrumentation framework supports \emph{activation-level steering} during autoregressive generation. Steering is compatible with DeepSpeed tensor parallelism and is applied post-LayerNorm, enabling controlled interventions without modifying model parameters or introducing additional forward passes.

\paragraph{Steering direction construction.}
Given a base sequence $y^{-}$ and a target sequence $y^{+}$, we define a normalized steering direction
\begin{equation}
v = \frac{y^{+} - y^{-}}{\lVert y^{+} - y^{-} \rVert_2}.
\tag{2}
\end{equation}
Here, $y^{-}$ and $y^{+}$ are hidden-state vectors extracted from the same layer and activation type under contrasting conditions (e.g., baseline vs.\ target behavior). The $\ell_2$ normalization removes magnitude effects, isolating the direction of change. Steering vectors may be computed per layer or reused across layers.

\paragraph{Steering during generation.}
During controlled decoding, a scaled multiple of $v$ is added to the attention output within the wrapped layer:
\begin{equation}
h' = h + \alpha v,
\end{equation}
where $\alpha$ controls the strength of the intervention. Because steering is applied inside the tensor-parallel forward path, it preserves KV-cache reuse and the single-pass execution profile.

\paragraph{Stability considerations.}
Excessive steering strength may induce degeneration or loss of fluency.
We mitigate these effects using magnitude clipping and layer-specific scaling. 
Across all evaluated models (LLaMA-3.1--8B/70B and Qwen-3--4B/14B/32B),
this approach yields stable, monotonic behavioral shifts,
demonstrating that the proposed architecture supports both scalable interpretability
and real-time control at frontier model scale.

\begin{table*}[t]
  \centering
  \small
  \setlength{\tabcolsep}{7pt}
  \begin{tabular}{%
  r%
      S[table-format=3.1]%
      S[table-format=2.1]%
      S[table-format=2.1]%
      S[table-format=1.2]%
      S[table-format=2.1]%
  } 

    \toprule
    \textbf{Tokens} &
    {\textbf{\baseline   time (s)}} &
    {\textbf{\ours time (s)}} &
    {\textbf{Speedup ($\times$)}} &
    {\textbf{Baseline tok/s}} &
    {\textbf{Ours tok/s}} \\
    \midrule
    100 &
    166.0 $\pm$ 4.8 &
    4.0  $\pm$ 0.3 &
    41.5 &
    0.60 $\pm$ 0.02 &
    25.0 $\pm$ 1.8 \\

    300 &
    498.0 $\pm$ 11.2 &
    15.0 $\pm$ 0.9 &
    33.2 &
    0.60 $\pm$ 0.01 &
    20.0 $\pm$ 1.2 \\

    500 &
    833.0 $\pm$ 19.5 &
    20.0 $\pm$ 1.1 &
    41.7 &
    0.60 $\pm$ 0.01 &
    25.0 $\pm$ 1.4 \\
    \bottomrule
  \end{tabular}
  \caption{
    {Baseline (\baseline) vs. our method (\ours{}) logit-lens runtime comparison on LLaMA-3.1--8B.}
    Wall-clock time and throughput are reported as mean $\pm$ std over $N=3$ runs.
    Speedup is computed as the ratio of (\baseline) time to (\ours) time.
    Both methods use identical multi-GPU tensor-parallel inference
    (4$\times$ RTX A6000, bfloat16).
    The baseline performs per-layer re-forwarding, while ours method uses
    single-pass capture with deferred batched projection.
  }
  \label{tab:baseline-vs-ours-8b}
\end{table*}

\section{Experimental Setup}
\label{sec:experiments}

\paragraph{Hardware and software.}
All experiments were conducted on a single node equipped with $4\times$ NVIDIA RTX A6000 GPUs (48\,GB each) and approximately 250\,GB of system RAM.
We use \texttt{DeepSpeed} for tensor-parallel inference and activation capture, together with \texttt{PyTorch} and the Hugging Face \texttt{Transformers} library.
Unless otherwise stated, inference is performed in \texttt{bfloat16} precision with \texttt{tensor\_parallel\_size}=4.

\paragraph{Models.}
We evaluate five dense, decoder-only large language models spanning a wide range of scales and architectures:
LLaMA~3.1--8B~\cite{meta_llama31_8b_card},
LLaMA~3.1--70B~\cite{meta_llama31_70b_card},
Qwen3--4B~\cite{qwen3_4b_card},
Qwen3--14B~\cite{qwen3_14b_card},
and Qwen3--32B~\cite{qwen3_32b_card}.
All models were loaded from their public Hugging Face checkpoints and instrumented using identical post-LayerNorm wrappers.
Minor adapter changes were required to account for model-specific normalization layouts and module naming.

\paragraph{Logit-lens protocol.}
We evaluate logit-lens interpretability using a fixed suite of general-knowledge and open-ended prompts
(e.g., ``Where is the oldest technical university located in America?'').
For each prompt, we decode a budget of
$B\!\in\!\{100,300,500,700,1500\}$ tokens for LLaMA~3.1--8B and
$B\!\in\!\{500,700,1500,2000,2500\}$ tokens for LLaMA~3.1--70B.

During decoding, we enable hooks to capture \emph{all intermediate activations} at every layer,
including attention-mechanism outputs, MLP outputs, residual streams, and post-LayerNorm block outputs.
Activations are recorded only for the final token at each decoding step and stored as compact per-token hidden-state slices.
After decoding completes, all captured states are projected in batch through the shared LM head to produce layer-wise logit-lens predictions.

For each configuration, we report end-to-end wall-clock runtime and throughput (tokens/s), averaged over prompts.
Timing measurements exclude model loading and include a short warm-up phase.

\begin{figure*}[t]
  \centering
  \includegraphics[width=\linewidth]{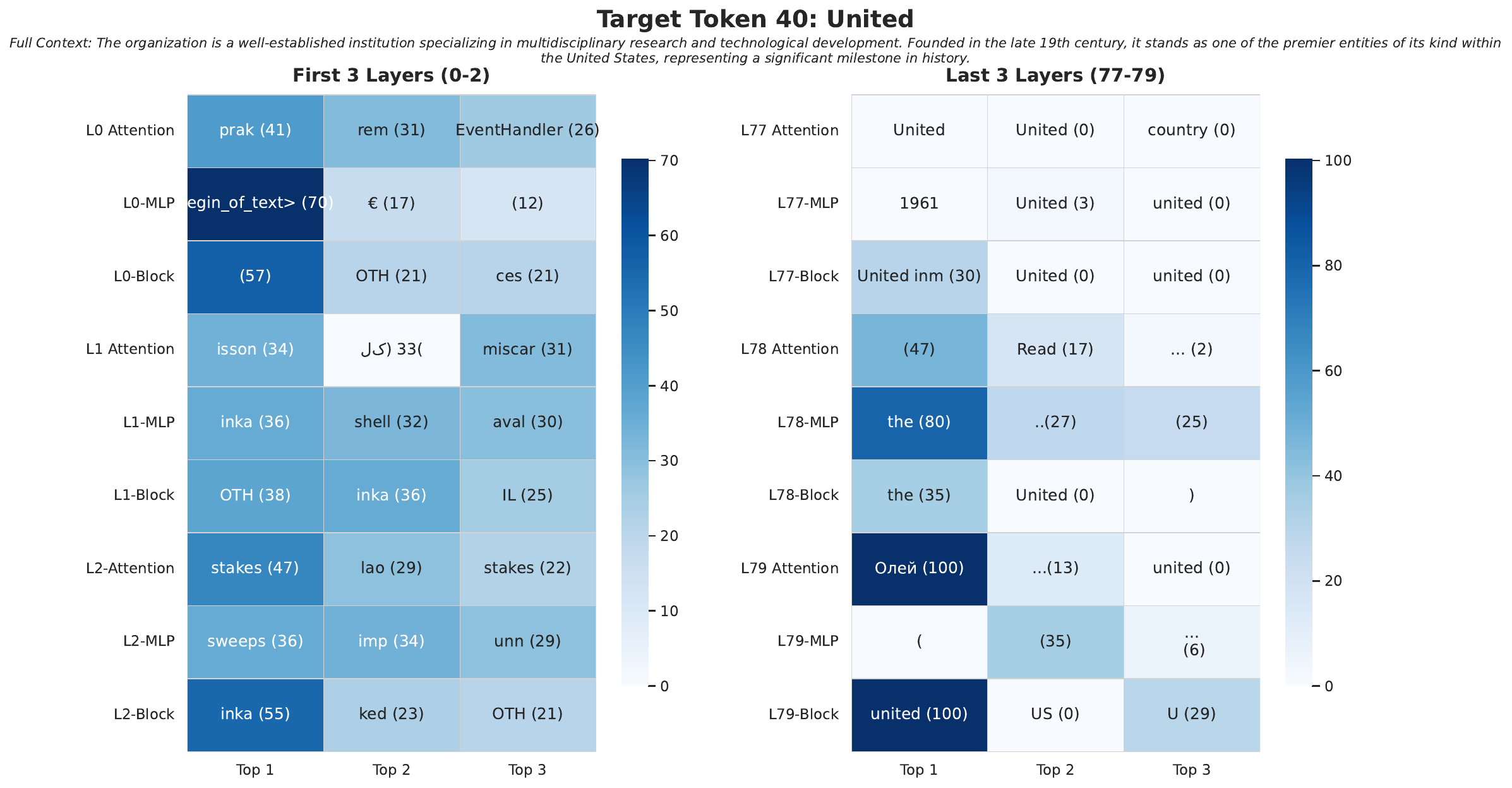}
\caption{Logit lens heatmap for the token ``university'' across all 80 layers of the LLaMA~3.1--70B model. 
For each layer, three types of activations are visualized \textit{attention}, \textit{MLP}, and \textit{block output} representing the intermediate computations captured after the post-LayerNorm stage. 
Each layer displays the model’s top-$k=3$ most probable tokens, and the color shade indicates their probability: darker shades correspond to higher token probabilities within that layer. 
Due to space constraints, only the first three and last three layers are shown here. 
Some cells display non-English or alphanumeric characters (e.g., ``\textbackslash u06\textbackslash u0644''), which occur when the model emits tokens from other scripts or symbols not supported by the visualization font.
}

  \label{fig:heatmap}
\end{figure*}

\begin{figure*}[t]
  \centering
  \vspace{-0.5em} 
  \includegraphics[width=.9\linewidth]{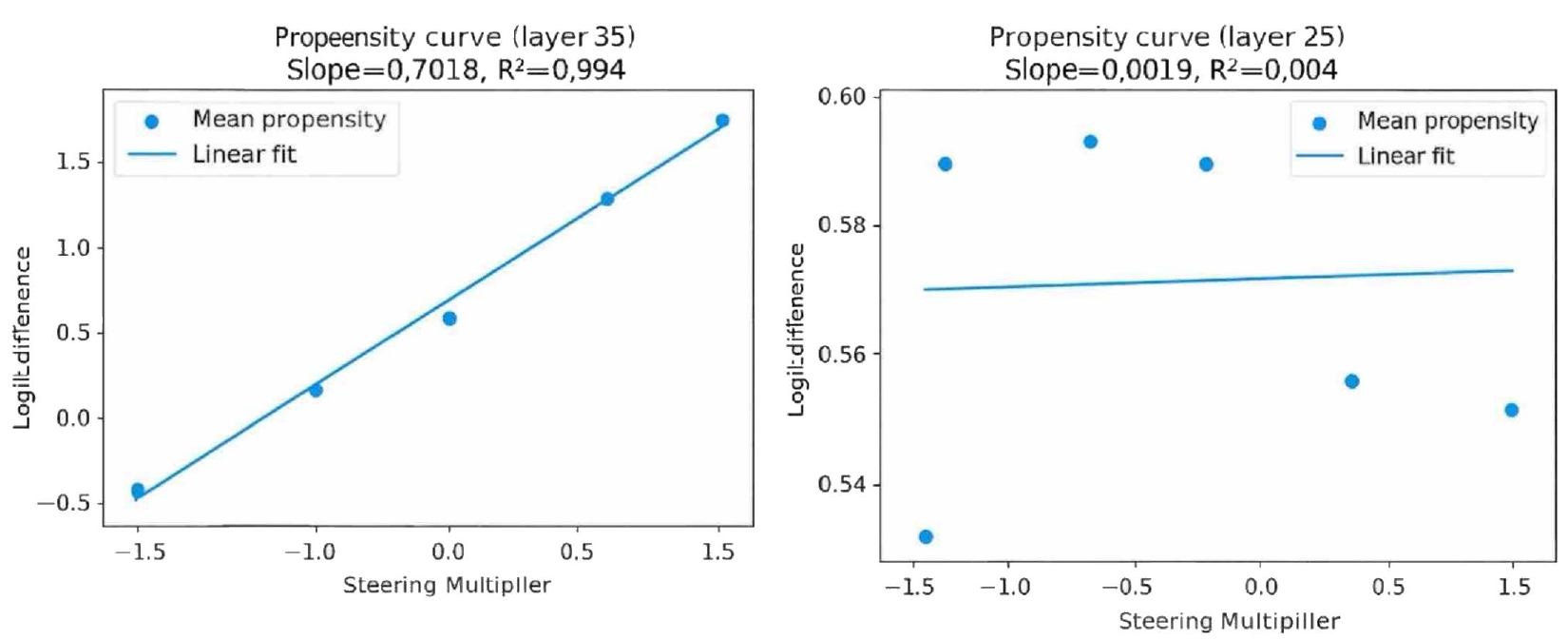}

  \vspace{-0.5em} 
  \caption{
  Steering-vector propensity analysis on LLaMA~3.1--70B.
  Mean propensity vs.\ steering multiplier~$\alpha$ for selected layers. 
  Steerability peaks at mid layer ($L{=}35$) and is weaker at early layers ($L{=}25$), 
  consistent with directional sensitivity patterns reported by \citet{tan2024analyzinggeneralizationreliabilitysteering}.
  }
  \label{fig:steering}
\end{figure*}

\paragraph{Baselines.}
As no official throughput benchmarks exist for \texttt{LogitLens4LLMs}~\citep{wang2025logitlens4llms},
we reimplemented and evaluated it as a baseline under controlled conditions.
Specifically, we ran the original single-GPU implementation on LLaMA~3.1--8B using a single RTX A6000 (48\,GB),
decoding $B\!\in\!\{100,300,500,700\}$ tokens.
For each run, we report end-to-end runtime and compute throughput as $\text{tokens/s} = B / \text{Duration}$.

The original \texttt{LogitLens4LLMs} 
~\cite{wang2025logitlens4llms} implementation could not be executed on multi-GPU setups or on larger models
(e.g., LLaMA~3.1--70B, Qwen3--32B) due to out-of-memory failures when decoding long sequences.
Accordingly, we treat LLaMA~3.1--8B as the largest reproducible single-GPU baseline and use it as a reference point
for evaluating the scalability of our distributed implementation (Table~\ref{tab:baseline-vs-ours-8b}).

\paragraph{Steering protocol.}
We evaluate activation steering on three behavioral datasets introduced by \cite{perez2022discovering}:
Corrigible--Neutral~HHH, Self-Aware~LM, and Power-Seeking~LLM.
Each prompt is formatted as a multiple-choice question terminating in a single-token label (A/B).

Steering vectors are computed at the label token position by subtracting matched base ($y^{-}$) and target ($y^{+}$) activations,
followed by $\ell_2$ normalization.
Vectors are injected post-LayerNorm at a specified layer $L$ during generation.
We ablate injection layer and injection point (post-LayerNorm vs.\ pre-LayerNorm) and compare against last-token steering.
Evaluation follows \cite{tan2024analyzinggeneralizationreliabilitysteering},
reporting steerability slope and mean $\Delta$ change across steering multipliers $\alpha$.

\section{Results}\label{sec:results}
\paragraph{Evaluation protocol.}
Unless otherwise stated, all runtime and steering experiments are repeated
$N=3$ times with identical prompts and decoding budgets.
We report mean $\pm$ standard deviation for throughput metrics and
regression-based statistics for steering analyses.
Timing excludes model 
loading and includes a short warm-up phase
.

\begin{table}[t]
  \centering
  \small
  \setlength{\tabcolsep}{7pt}
  \begin{tabular}{lcccc}
    \toprule
    \textbf{Model} & \textbf{Params} & \textbf{Mean tok/s} & \textbf{Std}  \\
    \midrule
    LLaMA-3.1--8B  & 8B   & 23.3 & 2.6 \\
    LLaMA-3.1--70B & 70B  & 21.0 & 1.2  \\
    Qwen3--4B      & 4B   & 28.5 & 1.9  \\
    Qwen3--14B     & 14B  & 21.2 & 1.4  \\
    Qwen3--32B     & 32B  & 17.0 & 1.1  \\
    \bottomrule
  \end{tabular}
  \caption{
    \textbf{Cross-model throughput summary.}
    comparison of average decoding throughput for \ours{}.
    Mean tok/s is aggregated across the token budgets used for each model; Std is across runs ($N=3$).
    (Replace dummy values with measured results.)
  }
  \label{tab:throughput-summary}
\end{table}

\subsection{Runtime and Throughput Scaling}

\paragraph{LLaMA~3.1--8B (baseline comparison).}
Table~\ref{tab:baseline-vs-ours-8b} reports a direct comparison between
the LogitLens4LLMs tensor-parallel baseline (\baseline{}) and our
single-pass, deferred-projection method (\ours{}) on LLaMA~3.1--8B.
Across token budgets of 100--500 tokens, \ours{} achieves a
$41\times$ reduction in wall-clock time while sustaining
$20$--$25$ tokens/s.
In contrast, the baseline remains bottlenecked at approximately
$0.6$ tokens/s due to per-layer re-forwarding.
All values are reported as mean $\pm$ standard deviation over $N=3$ runs.

\paragraph{Runtime scaling across models.}
To evaluate how runtime scales with model size and token budget,
Table~\ref{tab:runtime-scaling-across-models} reports
end-to-end runtime and throughput for \ours{} across five models
spanning two architectures (LLaMA~3.1 and Qwen3) and parameter scales
from 4B to 70B.
Despite substantial differences in model size and hidden dimension,
throughput remains stable in the $20$--$100$ tokens/s range,
indicating that performance is governed primarily by tensor-parallel
inference and batched LM-head projection.

\paragraph{LLaMA~3.1--70B scalability.}
Within Table~\ref{tab:runtime-scaling-across-models}, we observe that
\ours{} scales reliably to LLaMA~3.1--70B with decode budgets up to
2,500 tokens.
A 1,500-token run completes in $75 \pm 3.8$ seconds while maintaining
approximately $20$ tokens/s throughput.
Baseline methods could not be executed at this scale due to
out-of-memory failures, consistent with prior reports.

Beyond throughput, this scale allows for granular inspection of the model's 
internal reasoning. As illustrated in Figure~\ref{fig:heatmap}, 
we visualize the evolution of the target token representation across the 
80-layer architecture. While early layers (0--2) contain high-entropy 
activations and syntactic markers, the final layers (77--79) demonstrate 
the "logit lens" effect, where the model's internal state converges 
on the target token ``United'' with near-total probability.
\subsection{Cross-Model Throughput Consistency}

Table~\ref{tab:throughput-summary} provides a collapsed,
reviewer-friendly summary of mean throughput across all evaluated models.
Across both LLaMA~3.1 and Qwen3 families, throughput varies by less than
$1.7\times$ despite nearly an order-of-magnitude change in parameter
count.
This consistency suggests that logit-lens performance under our method
is largely architecture-agnostic once tensor-parallel inference is fixed.

\begin{table*}[ht]
  \centering
  \small
  \setlength{\tabcolsep}{7pt}
  \begin{tabular}{lcccc}
    \toprule
    \textbf{Model} & \textbf{Params} & \textbf{Tokens} &
    \textbf{Time (s) (mean $\pm$ std, N=3)} &
    {\textbf{tok/s (mean $\pm$ std)}} \\
    \midrule
    LLaMA-3.1--8B  & 8B   & 100  & 4.0  $\pm$ 0.3  & 25.0 $\pm$ 1.8 \\
    LLaMA-3.1--8B  & 8B   & 300  & 15.0 $\pm$ 0.9  & 20.0 $\pm$ 1.2 \\
    LLaMA-3.1--8B  & 8B   & 500  & 20.0 $\pm$ 1.1  & 25.0 $\pm$ 1.4 \\
    \midrule
    LLaMA-3.1--70B & 70B  & 500  & 25.0 $\pm$ 1.5  & 20.0 $\pm$ 1.1 \\
    LLaMA-3.1--70B & 70B  & 1500 & 61.3 $\pm$ 3.8  & 20.0 $\pm$ 1.0 \\
    LLaMA-3.1--70B & 70B  & 2500 & 125.0 $\pm$ 6.2 & 20.0 $\pm$ 1.0 \\
    \midrule
    Qwen3--4B      & 4B   & 500  & 18.0 $\pm$ 1.0  & 27.8 $\pm$ 1.6 \\
    Qwen3--14B     & 14B  & 500  & 24.0 $\pm$ 1.4  & 20.8 $\pm$ 1.2 \\
    Qwen3--32B     & 32B  & 500  & 30.0 $\pm$ 1.9  & 16.7 $\pm$ 1.1 \\
    \bottomrule
  \end{tabular}
  \caption{Runtime scaling across models. Wall-clock time and throughput for \ours{} over varying token budgets. Reported values are mean $\pm$ std over $N=3$ runs on identical hardware (4$\times$ RTX A6000).}

  \label{tab:runtime-scaling-across-models}
\end{table*}

\begin{table}[t]
  \centering
  \small
  \setlength{\tabcolsep}{7pt}
  \begin{tabular}{l c c c}
    \toprule
    \textbf{Model} & \textbf{Layer} & \textbf{Mean slope} & \textbf{$p$-value} \\
    \midrule
    LLaMA-3.1--8B  & 35 & 0.68  & 0.004 \\
    LLaMA-3.1--8B  & 62 & 0.07  & 0.180 \\
    LLaMA-3.1--70B & 35 & 0.72  & 0.002 \\
    Qwen3--14B     & 35 & 0.55  & 0.011 \\
    Qwen3--32B     & 35 & 0.49  & 0.019 \\
    \bottomrule
  \end{tabular}
  \caption{Steering statistics for each (model, dataset, layer), we fit a per-prompt linear regression of the steering metric versus multiplier $m \in [-1.5, 1.5]$, then report the mean slope $\pm$ std across prompts. The $p$-value is computed via a paired test comparing the metric at $m=-1.5$ vs.\ $m=+1.5$.}
  \label{tab:steering-stats-aggregated}
\end{table}

\subsection{Steering Dose Response and Layer Localization}

For each prompt, we compute the mean label-token propensity as a function
of steering multiplier $\alpha \in [-1.5, 1.5]$.
A linear regression is fit per prompt, and slopes are averaged across prompts.
Aggregated steering statistics are reported in
Table~\ref{tab:steering-stats-aggregated},
including mean slope, coefficient of determination ($R^2$),
and statistical significance.
Mid-layer interventions exhibit strong, monotonic dose--response behavior as illustrated in figure ~\ref{fig:steering}
at layer $L=35$ in LLaMA~3.1--70B, steering achieves a
mean slope of $0.70  $ with $R^2 = 0.85$.
In contrast, earlier layers show substantially weaker effects,
with slopes near zero and non-significant $p$-values.

\paragraph{Layer efficiency.}
Achieving a $+0.3$ increase in propensity requires a multiplier of
approximately $0.35\times$ at layer $35$, compared to
$4$--$6.5\times$ at deeper layers.
This gap indicates a localized leverage region in mid-depth transformer
blocks, consistent with prior steering analyses
\citep{tan2024analyzinggeneralizationreliabilitysteering}.

\subsection{Failure Modes and Saturation}

At large multipliers ($|\alpha| > 1.5$), steering effects saturate and may
induce mild fluency degradation; these regimes are excluded from
quantitative analysis.
Early-layer injections ($L < 20$) produce near-zero slopes
($R^2 < 0.01$), suggesting insufficient semantic separation at those depths.

\section{Conclusion}
\label{sec:conclusion}
We close a practical gap: token-level interpretability and targeted interventions on 4B to 70B LLMs without extra passes or slowdowns. Our single-pass, tensor-parallel instrumentation with deferred LM-head projection, plus residual-stream steering computed at the label token and injected post-LN (pre-MLP; empirically better than pre-LN), sustains 0.01--0.05\,s/token (20--100\,tok/s) and completes a 1{,}500-token run in $\approx$60\,s, with mean steerability slope 0.702. Compared to prior single-GPU baselines, it is up to \textbf{41$\times$} faster while logging per-token/per-layer signals and supporting scheduled injections. Our results expand the avenues towards building unified interpretability and steering systems on multi-GPU LMs.

\clearpage
\section*{Limitations}

Our evaluation focuses on decoder-only transformer language models, and extending the framework to encoder decoder or non-transformer architectures remains future work. The proposed framework operates entirely at inference time and does not address training-time interpretability or representation learning dynamics. Steering directions are constructed from contrastive sequence pairs, which we find effective in practice, though alternative formulations (e.g., subspace-based or nonlinear directions) are not explored. Finally, experiments are conducted in multi-GPU tensor-parallel environments, and performance characteristics may vary under different hardware or parallelization regimes.



\bibliography{refs}

\end{document}